\documentclass[letterpaper]{article} 
\usepackage[preprint]{aaai2027}  
\usepackage[hyphens]{url}  
\usepackage{graphicx} 
\usepackage{natbib}  
\usepackage{caption} 
\usepackage{algorithm}
\usepackage{algorithmic}

\usepackage{newfloat}
\usepackage{listings}
\DeclareCaptionStyle{ruled}{labelfont=normalfont,labelsep=colon,strut=off} 
\floatstyle{ruled}
\newfloat{listing}{tb}{lst}{}
\floatname{listing}{Listing}

\usepackage{booktabs}

\usepackage{amsmath}
\usepackage{amssymb}
\usepackage{xcolor}
\newcommand{\method}{A4A}

\begin{document}


\title{$\mathcal{A}\textit{4}\mathcal{A}$:
Cross-Embodiment Transfer of Action-Oriented 4D Affordances\\
from Human Demonstrations}
\author{
Yifan Han\textsuperscript{1,*},
Litao Liu\textsuperscript{2,*},
Yuqi Gu\textsuperscript{1},
Ye Lu\textsuperscript{1},
Hanqing Wang\textsuperscript{4},\\
Sidney Wai\textsuperscript{2},
Ishaan Myrie\textsuperscript{2},
Qi Zhang\textsuperscript{5},
Jingjin Yu\textsuperscript{2,\dag},
Gen Li\textsuperscript{3,\dag}
}
\affiliations{
\textsuperscript{1}Shanghai Jiao Tong University \\
\textsuperscript{2}Rutgers University--New Brunswick\\
\textsuperscript{3}Nanyang Technological University \\
\textsuperscript{4}The Hong Kong University of Science and Technology (GZ)\\
\textsuperscript{5}Shanghai AI Laboratory\\
\textsuperscript{*}Equal contribution. \quad
\textsuperscript{\dag}Corresponding author. \\
\textcolor{blue}{\url{https://ru-arcl.github.io/a4a/}}%
}
\makeatletter
\g@addto@macro\@maketitle{%
  \par
  \centering
      \includegraphics[width=\linewidth]{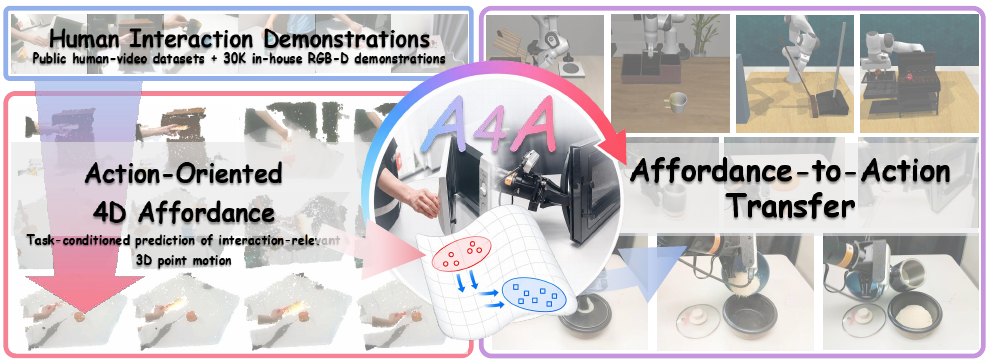}
            \captionof{figure}
            {\textbf{What should robots learn from human demonstrations?} We identify the task-conditioned future motion of interaction-relevant 3D geometry as a transferable signal across embodiments. \method{} uses these action-oriented 4D affordances to pretrain VLA policies before adapting them to robot control.}
    \label{fig:teaser}
  \par
}
\makeatother

\newcommand{\authornote}[1]{%
  \begingroup
  \renewcommand{\thefootnote}{}%
  \footnotetext{#1}%
  \endgroup
}

\maketitle

\authornote{%
Corresponding Author: Gen Li (\mbox{gen.li@ntu.edu.sg}), Jingjin Yu (\mbox{jingjin.yu@cs.rutgers.edu}).
Project Leader: Yifan Han (\mbox{hanyifan2024@ia.ac.cn}), Litao Liu (\mbox{litao.liu@rutgers.edu}).
Project page: \textcolor{blue}{\url{https://ru-arcl.github.io/a4a/}}.%
}

\begin{abstract}
Human demonstrations contain rich manipulation knowledge, but it remains unclear what information can be transferred effectively to robot control. Existing affordance representations are typically formulated as 2D masks, 3D regions, contact points, or actionability scores, and therefore primarily identify where interaction may occur. However, effective manipulation also requires modeling how interaction-relevant geometry evolves during task execution. To bridge this gap, we introduce \emph{action-oriented 4D affordances}, which represent the language-conditioned future trajectories of interaction-relevant 3D points. 
These trajectories capture task-conditioned geometric evolution rather than embodiment-specific actions, enabling transferable interaction priors across humans and robots.
Based on this representation, we construct a large-scale action-oriented 4D affordance dataset from existing human--object interaction video data and complementary RGB-D demonstrations, and introduce \method{}, an affordance-to-action framework that uses 4D affordance trajectory prediction to pretrain robot policies before manipulation finetuning. Experiments in both simulation and the real world validate the effectiveness of \method{}, showing that pretraining with action-oriented 4D affordance data consistently improves the manipulation performance of diverse VLA policies. These results establish action-oriented 4D affordances as an effective cross-embodiment representation for transferring manipulation knowledge from human demonstrations to robot control.
\end{abstract}

\section{Introduction}



Human videos provide a scalable and diverse source of supervision for robot learning, capturing rich object interactions across tasks, environments, and human behaviors. Despite their richness, it remains unclear which information in human videos is most useful for learning robot control. For example, a video of a person opening a drawer offers multiple candidate supervision signals, including visual appearance, human motion, object-centric interaction cues, and task semantics. Many of these signals are either tied to human embodiment or provide only indirect supervision for control. 
A robot should neither imitate a human arm trajectory nor infer how to complete a task from object identity alone. The central question is, therefore, \textit{which signal in human videos can directly guide control while remaining transferable across embodiments.}


In robotics and computer vision, affordance is commonly represented as 2D masks~\cite{li2025learning}, 3D region~\cite{wu2025open,fang2026saga}, contact points~\cite{yuan2024robopoint}, actionable object part~\cite{robo-abc}, or actionability score~\cite{mo2021where2act}. These representations provide useful spatial grounding by identifying where an interaction may occur~\cite{vrb}. Yet manipulation requires more than localizing an actionable region. A drawer handle does not determine the pulling displacement and a cup rim does not define the motion required for pouring. Static affordance representations therefore describe \emph{where} to interact, while providing limited guidance on \emph{how} the interaction should unfold.


We argue that the transferable signal in human demonstrations should be operational rather than merely spatial. What robots need from human data is the task-conditioned motion of interaction-relevant 3D geometry. When a person pulls a drawer, local points on the handle and drawer translate outward. When a person opens a cap, local points rotate around an axis. When a person presses a button, points near the contact region move along a short normal direction. These motions are visible in human demonstrations and describe the physical state transition that a robot action must realize.

Our formulation builds on a simple geometric observation. Over a short horizon, interaction-relevant 3D point motion in human demonstrations and robot end-effector motion share a local geometric structure: both can often be approximated by rigid or quasi-rigid transformations in $SE(3)$. Although the exact motion depends on embodiment, contact condition, and coordinate frame, these shared short-horizon motion patterns provide transferable interaction priors across humans and robots. Thus, while human and robot motor commands live in different embodiment-specific action spaces, interaction-relevant 3D point motion in human demonstrations and robot end-effector displacement provide two geometric descriptions of the same underlying manipulation transition. This observation motivates a transferable representation for learning
from humans: neither human joint motion nor merely a static affordance
mask, but the future motion of interaction-relevant 3D geometry.


We call this representation an \emph{action-oriented 4D affordance}, defined as a language-conditioned trajectory field over 3D query points. The initial query points ground the interaction in the scene, while their future trajectories specify the operation. Unlike 2D or 3D affordance masks, the additional temporal dimension changes the nature of the supervision: the objective is no longer merely ``where can the robot act?'' but ``how should the interaction geometry evolve over time?'' As a result, a 4D affordance serves as an action-oriented representation that connects perception with control rather than functioning solely as a perceptual descriptor.

Building upon this representation, we introduce \method{}, a cross-embodiment transfer framework that uses action-oriented 4D affordances to initialize robot policies from human demonstrations. As illustrated in Figure~\ref{fig:teaser}, \method{} extracts identity-preserving 3D point trajectories from human videos and uses them as an intermediate prediction space for VLA pretraining. During robot adaptation, the learned internal representation is retained, while the embodiment-specific state and action interfaces of the base policy are restored. This design transfers human-observed interaction geometry to robot control across different VLA action-generation paradigms.
Our contributions are summarized as follows:
\begin{itemize}
    \item We introduce \emph{action-oriented 4D affordances}, a task-conditioned representation of the future evolution of interaction-relevant 3D geometry captured from human demonstrations. Unlike static affordance localization, they characterize how an interaction should unfold and provide a transferable, action-aligned supervision signal across embodiments.

    \item We propose \method{}, a cross-embodiment affordance-to-action transfer framework that uses 4D point trajectories as the pretraining prediction space of VLA policies and adapts the learned representation to robot control through each policy's native state and action interfaces. The framework is instantiated according to the native action-generation paradigm of each base policy.

    \item We evaluate \method{} across five VLA policy families, simulated manipulation benchmarks, and real-world tasks. The results demonstrate consistent downstream gains across autoregressive, regression, diffusion, and flow-matching policies, strong RGB-only performance against methods that explicitly use 3D or 4D observations, and more effective transfer than generic scene-wide 4D pretraining in an architecture-controlled comparison.
\end{itemize}


\section{Related Work}

\paragraph{VLA and Affordance Learning.}  
Affordance learning provides an important inductive bias for language-conditioned manipulation by grounding task semantics in functional interaction structure. Existing methods either expose affordance as explicit policy input, such as affordance plans, contact regions, or post-contact motion cues~\citep{rt-aff, zhang2024affordance, wu2025afforddp}, or internalize it as latent supervision within VLA-style policies~\citep{li2025coa, kong2026affordvla, yu2026affordancevla, liu2026affordance2action}. These works improve the connection between perception, language, and action, but most affordance representations remain primarily spatial: they identify objects, regions, contacts, or action primitives. In contrast, our work treats affordance as a 4D operational signal. Rather than only predicting where interaction should occur, we model how interaction-relevant 3D geometry should move over time.


\paragraph{Learning Robot Manipulation from Human Demonstrations.}  Human demonstrations offer a scalable source of manipulation knowledge and can reduce the need for costly robot demonstrations. Prior work leverages human videos either by jointly training with robot data~\citep{myers2024policy, lee2025tracegen, bharadhwaj2024track2act}, or by learning transferable priors from human demonstrations before downstream robot adaptation~\citep{wen2023any, niu2025pre}. Recent foundation-model-based approaches further extract execution-relevant information such as subtask structure, key states, or task constraints from human behavior~\citep{wang2025vlm, huang2024rekep}. While effective, these methods often rely on high-level abstractions or require robot data to bridge the embodiment gap. Our work instead extracts a lower-level transferable signal from human videos: task-conditioned 4D point motion that directly describes the geometric transition induced by manipulation.


\paragraph{Embodiment-Agnostic Representations for Manipulation.}  The embodiment gap makes direct imitation of human motion unsuitable for robot control, motivating representations that describe manipulation independently of the acting body. 2D point tracks and flow-like signals provide one such interface~\citep{bharadhwaj2024track2act, wen2023any, xu2024flow}, but they lack metric 3D structure and are sensitive to viewpoint ambiguity. Structured 3D and 4D motion representations address this limitation by modeling point trajectories, scene flow, or physically grounded dynamics in space and time~\citep{yuan2024general, niu2025pre, han2026bridgeact, pointworld}. Our work follows this direction but focuses on interaction-relevant rather than generic scene motion. We define action-oriented 4D affordances as the future trajectories of task-relevant 3D points, providing a geometrically grounded representation aligned with manipulation.


\section{Method}

\begin{figure*}[!t]
    \centering
    \includegraphics[width=\textwidth]{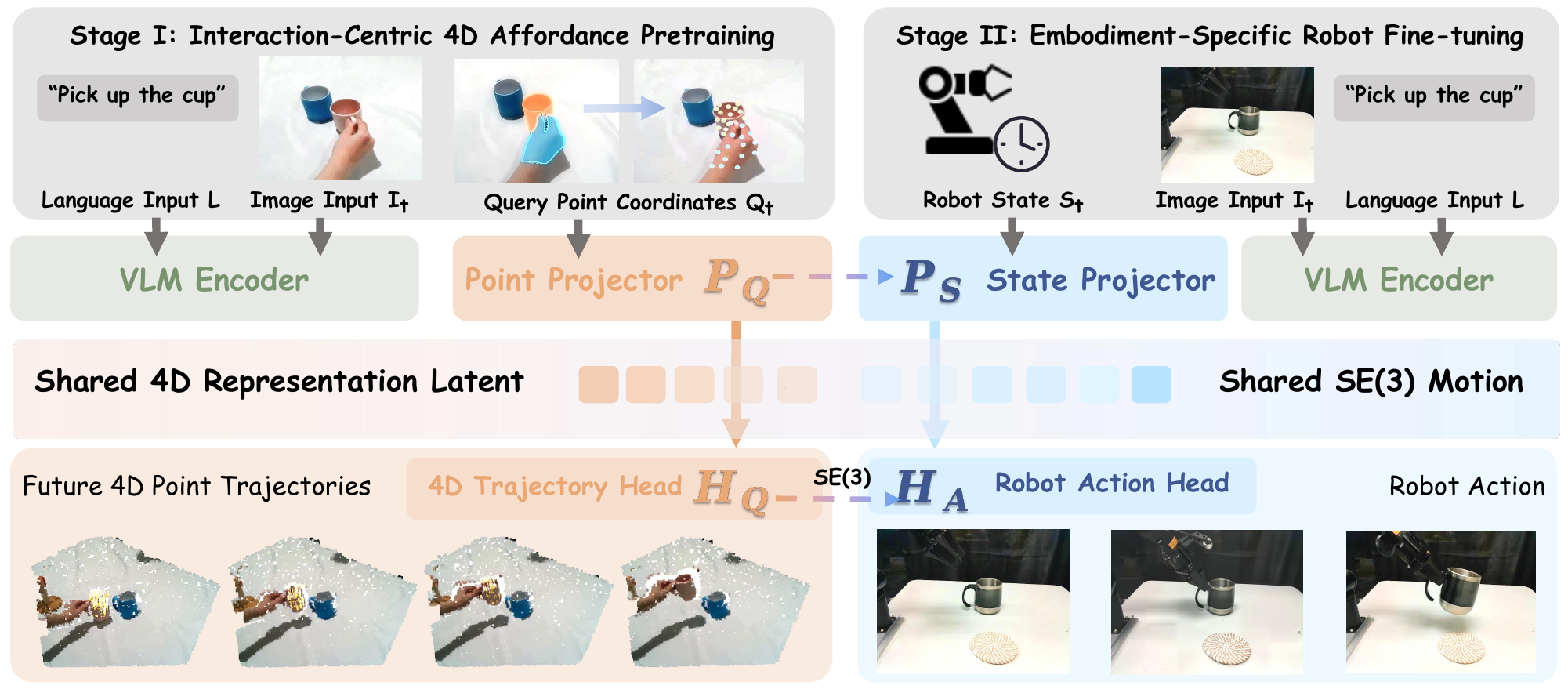}
        \caption{
            \textbf{Affordance-to-action representation transfer.} Guided by the short-horizon geometric alignment between interaction-relevant 4D point motion and robot end-effector motion, \method{} pretrains a vision--language policy to predict future 3D point trajectories and transfers the learned representation to robot control through the policy's native state and action interfaces.}
    \label{fig:pipeline} 
    \vspace{-5mm}
\end{figure*}

\method{} is a cross-embodiment pretraining framework that transfers
manipulation knowledge from human demonstrations to robot control
through action-oriented 4D affordances. As illustrated in
Figure~\ref{fig:pipeline}, we first motivate this representation through
the short-horizon correspondence between interaction-relevant 4D point
motion and robot end-effector displacement, and then construct a
large-scale 4D affordance corpus from human demonstrations. Building on
this representation and corpus, we adopt a two-stage training strategy:
the policy is first pretrained to predict task-conditioned future
trajectories of interaction-relevant 3D points, and is subsequently
adapted to robot control by restoring its native state and action
interfaces while retaining and fine-tuning the pretrained
manipulation-oriented representation.

\subsection{Geometric Premise for Cross-Embodiment Transfer}

Our central premise is that the transferable quantity across embodiments is the geometric effect of manipulation, rather than the embodiment-specific command that produces it. \noindent\textbf{Over a short horizon, end-effector motion and the 4D motion of interaction-relevant local points can be viewed as two geometric representations of the same manipulation transition.} This correspondence is exact for points rigidly associated with the gripper, tool, or grasped object. Let $G_t\in SE(3)$ denote the transformation from the robot end-effector frame to a fixed reference frame. For a point with fixed homogeneous coordinate $\bar{\mathbf{q}}_i^{E}$ in the end-effector frame, its position at time $t$ is
\begin{equation}
\bar{\mathbf{q}}_i^t
=
G_t\bar{\mathbf{q}}_i^{E}.
\label{eq:point_from_ee_pose}
\end{equation}
Thus, the 3D geometry rigidly attached to the gripper, tool, or grasped object is fully determined by the end-effector pose. The end-effector displacement between two consecutive timesteps is
\begin{equation}
A_t
=
G_{t+1}G_t^{-1}
\in SE(3).
\label{eq:ee_displacement}
\end{equation}
The same displacement directly induces the motion of every attached point:
\begin{equation}
\bar{\mathbf{q}}_i^{t+1}
=
A_t\bar{\mathbf{q}}_i^t.
\label{eq:point_action_correspondence}
\end{equation}
End-effector displacement and the induced 4D motion of attached points therefore represent the same short-horizon spatial transition, making 4D point motion a natural intermediate representation for robot actions. For broader interaction-relevant geometry in human videos, including points on the manipulated object, tool, or contacted object part, we use this correspondence as a local short-horizon approximation: over the prediction horizon, their motion can often be approximated by a rigid or quasi-rigid transformation in $SE(3)$. Human videos expose this interaction-relevant geometric evolution without providing robot control commands; we thus use task-conditioned 4D trajectories as pretraining supervision to learn transferable priors, which are later adapted to embodiment-specific control during finetuning. Unlike static affordances that specify \emph{where} to act, 4D affordances specify \emph{how} the relevant geometry should move.

\subsection{4D Affordance Data Construction}
\label{sec:data_construction}

We construct an action-oriented 4D affordance dataset from existing human--object interaction data~\cite{liu2022hoi4d,damen2020epic} and complementary in-house RGB-D demonstrations. It contains over 80K interaction clips, including approximately 30K collected in-house. The existing data covers common manipulation behaviors such as opening, closing, picking up, placing, pushing, and pulling, while our recordings extend the coverage to underrepresented operations including pouring, cutting, hanging, sweeping, and lid removal. Detailed statistics, affordance categories, and qualitative visualizations are provided in Appendix B.

Samples from existing datasets are processed using their available interaction annotations, while the in-house demonstrations are captured with an Intel RealSense RGB-D camera. We refine the raw depth maps using LingBot-Depth~\cite{tan2026masked}. Given text prompts derived from the task instruction, GroundingDINO~\cite{liu2024grounding} localizes the manipulated object, tool, or contacted object part, and SAM~2~\cite{ravi2025sam} produces the corresponding mask from the detected bounding box or manually specified point prompts. Query points sampled from the selected region are tracked across frames using CoTracker3~\cite{karaev2025cotracker3} and back-projected into 3D using the refined depth maps and camera intrinsics, producing identity-preserving 3D point trajectories.

Each demonstration is represented in a unified action-oriented 4D affordance format:
\begin{equation}
\mathcal{S}=\left(I^{0:H},l,Q_{\mathrm{int}}^{0:H}\right),
\label{eq:4d_affordance_sample}
\end{equation}
where $I^{0:H}$ denotes the visual observation sequence, $l$ is the task instruction, and $Q_{\mathrm{int}}^{0:H}$ contains the 3D trajectories of query points sampled from the manipulated object, tool, or contacted object part. The initial points ground the interaction-relevant geometry, while their future positions encode the task-conditioned geometric transition.

\subsection{Affordance-to-Action Representation Transfer}
\label{sec:affordance_transfer}

\method{} adopts a two-stage training strategy based on prediction-space substitution. For each base VLA, we reuse its native vision--language conditioning stack and the internal module responsible for action generation. During 4D affordance pretraining, query-point states replace robot states and future point motion replaces the robot-action target. During robot finetuning, the original proprioceptive and action interfaces are restored, while the parameters learned from 4D prediction initialize the downstream policy. The same internal representation is therefore used to model both interaction-centric point motion and embodiment-specific robot actions, without introducing an additional temporal network. For OpenVLA, whose native action space is discretized and autoregressive, we instantiate the same transfer principle through a tokenized 4D residual formulation that reuses its language-model decoder and output head; implementation details are provided in Appendix A.

\paragraph{Interaction-Centric 4D Pretraining.}

Given the current image $I_t$, task instruction $l$, and interaction-relevant query points $Q_t$, the native vision--language front end of the base policy produces multimodal context features. A lightweight point projector $P_Q$ maps the query-point coordinates into the model's state-token space. The visual, language, and point-state representations are then processed by the shared latent module $F_\theta$, and a point prediction interface $H_Q$ maps the resulting representation to future 4D point motion:
\begin{equation}
\begin{aligned}
\mathbf{v}_t
&=
E_{\mathrm{VLM}}(I_t,l),\\
\mathbf{z}_t^Q
&=
F_{\theta}
\left(
\mathbf{v}_{\leq t},
P_Q(Q_{\leq t})
\right),\\
\widehat{\Delta Q}_{t+1:t+H}
&=
H_Q(\mathbf{z}_t^Q).
\end{aligned}
\label{eq:4d_affordance_pretraining}
\end{equation}
Here, $E_{\mathrm{VLM}}$ denotes the native visual--language conditioning modules, and $F_\theta$ denotes the internal module shared between 4D pretraining and robot-action generation; depending on the base policy, it may be a language-model decoder, multimodal Transformer, or action expert. The point projector $P_Q$ maps 3D coordinates to the policy's input representation, while $H_Q$ expresses its native prediction pathway in the 4D point space. In our implementation, the point target is parameterized as a residual over a constant-velocity prediction, as detailed in Appendix A.

Unlike generic scene-wide point prediction, our supervision is restricted to points on the manipulated object, tool, or contacted object part. This interaction-centric target directs the shared representation toward the task-relevant geometric transition induced by the demonstrated operation, rather than motion from unrelated scene regions.
\begin{figure*}[!t]
    \centering
    \includegraphics[width=\textwidth]{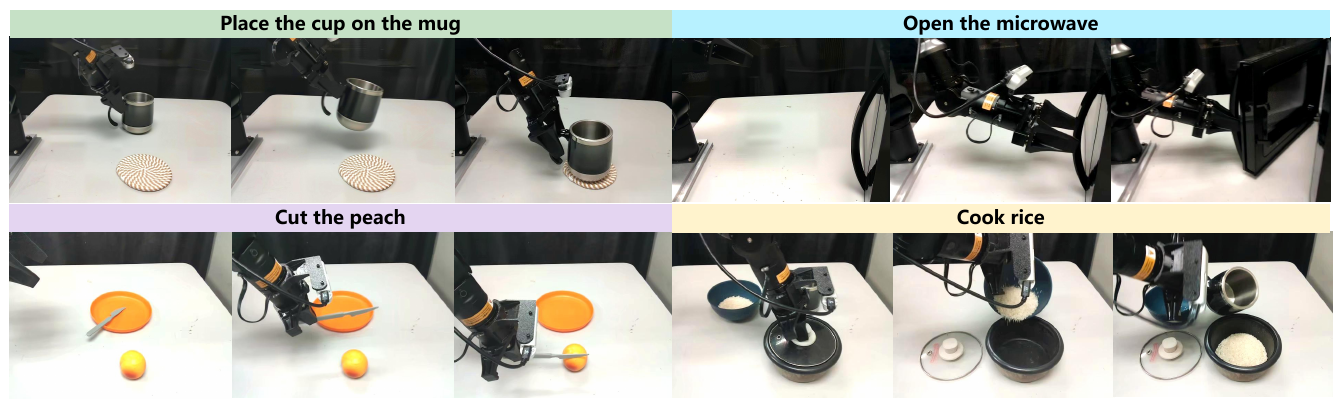}
        \caption{
            \textbf{Real-world manipulation.} Representative \method{} rollouts on placing a cup on a mug, opening a microwave, cutting a peach, and cooking rice.}
    \label{fig:piper_snapshots}
    \vspace{-5mm}
\end{figure*}

We retain the native objective family of each base policy during pretraining and apply it to the valid 4D point targets:
\begin{equation}
\mathcal{L}_{\mathrm{aff}}
=
\mathcal{J}_{\mathrm{base}}
\left(
\widehat{\Delta Q}_{t+1:t+H},
\Delta Q_{t+1:t+H};
M
\right),
\label{eq:affordance_loss}
\end{equation}
where $M$ masks point trajectories affected by invalid depth, occlusion, or tracking failure. The function $\mathcal{J}_{\mathrm{base}}$ denotes the prediction objective native to the selected VLA, applied to point motion rather than robot actions. Consequently, the shared latent module is pretrained through the same prediction mechanism that it later uses for action generation, avoiding an objective mismatch between the two stages.

\paragraph{Robot Action Finetuning.}

To adapt the pretrained representation to robot control, we replace the point projector $P_Q$ with a proprioceptive-state projector $P_S$ at the corresponding state interface. The 4D prediction interface $H_Q$ is replaced by the robot-action interface $H_A$ of the downstream policy. The native vision--language conditioning stack and shared latent module are initialized from 4D affordance pretraining and jointly finetuned on robot demonstrations:
\begin{equation}
\begin{aligned}
\mathbf{v}_t^R
&=
E_{\mathrm{VLM}}(o_t,l),\\
\mathbf{z}_t^S
&=
F_{\theta}
\left(
\mathbf{v}_{\leq t}^R,
P_S(s_{\leq t})
\right),\\
\widehat{\mathbf{a}}_{t:t+K-1}
&=
H_A(\mathbf{z}_t^S).
\end{aligned}
\label{eq:robot_action_finetuning}
\end{equation}
Here, $o_t$ and $s_t$ denote the robot observation and proprioceptive state, respectively, and $\widehat{\mathbf{a}}_{t:t+K-1}$ denotes the predicted action chunk. The action interface $H_A$ follows the native action representation and training objective of the selected VLA. Thus, robot finetuning restores the original control formulation of the base policy rather than introducing a separate action-prediction architecture.

The transition from 4D affordance pretraining to robot control is therefore localized to the external state and prediction interfaces:
\begin{equation}
P_Q \rightarrow P_S,
\qquad
H_Q \rightarrow H_A.
\label{eq:interface_substitution}
\end{equation}
The vision--language conditioning modules and the shared latent parameters $F_\theta$ are preserved across the two stages. Under the short-horizon geometric correspondence established above, the two stages represent the same manipulation transition in different spaces: 4D point motion provides a human-observable geometric representation, while robot actions provide its embodiment-specific realization. Sharing the latent module and objective family enables the interaction prior learned from human demonstrations to transfer to downstream control.

\section{Experiments}

Our experiments address the following questions:

\begin{itemize}
    \item How effectively does \method{} improve VLA policies, and are its gains consistent across diverse action-generation paradigms?
    \item How does an RGB policy initialized with 4D affordance pretraining compare with methods that use explicit 3D representations, and how much of its gain comes from our pretraining data rather than the architectural differences?
    \item How well does the learned 4D prior transfer to real-world manipulation across different VLA policies?
\end{itemize}

\subsection{Experimental Setup}

\begin{table*}[!t]
\centering
\small
\setlength{\tabcolsep}{3pt}
\begin{tabular*}{\textwidth}{@{\extracolsep{\fill}}l ccc ccc ccc ccc ccc@{}}
\toprule
& \multicolumn{3}{c}{Octo}
& \multicolumn{3}{c}{OpenVLA}
& \multicolumn{3}{c}{OpenVLA-OFT}
& \multicolumn{3}{c}{$\pi_0$}
& \multicolumn{3}{c}{$\pi_{0.5}$} \\
\cmidrule(lr){2-4}
\cmidrule(lr){5-7}
\cmidrule(lr){8-10}
\cmidrule(lr){11-13}
\cmidrule(lr){14-16}

Object
& Base & \method{} & $\Delta$
& Base & \method{} & $\Delta$
& Base & \method{} & $\Delta$
& Base & \method{} & $\Delta$
& Base & \method{} & $\Delta$ \\
\midrule

Alphabet soup
& 36 & 64 & +28
& 76 & 88 & +12
& 100 & 100 & 0
& 76 & 88 & +12
& 100 & 100 & 0 \\

Cream cheese
& 20 & 56 & +36
& 68 & 78 & +10
& 98 & 98 & 0
& 86 & 92 & +6
& 90 & 94 & +4 \\

Salad dressing
& 30 & 68 & +38
& 84 & 92 & +8
& 98 & 98 & 0
& 86 & 94 & +8
& 100 & 98 & -2 \\

BBQ sauce
& 16 & 36 & +20
& 42 & 54 & +12
& 98 & 100 & +2
& 56 & 78 & +22
& 84 & 92 & +8 \\

Ketchup
& 12 & 48 & +36
& 74 & 86 & +12
& 100 & 98 & -2
& 88 & 92 & +4
& 100 & 100 & 0 \\

Tomato sauce
& 38 & 58 & +20
& 58 & 74 & +16
& 100 & 100 & 0
& 72 & 86 & +14
& 96 & 98 & +2 \\

Butter
& 18 & 38 & +20
& 62 & 68 & +6
& 98 & 100 & +2
& 88 & 96 & +8
& 92 & 96 & +4 \\

Milk
& 14 & 56 & +42
& 78 & 84 & +6
& 100 & 100 & 0
& 62 & 74 & +12
& 84 & 88 & +4 \\

Chocolate pudding
& 54 & 72 & +18
& 56 & 62 & +6
& 88 & 94 & +6
& 90 & 96 & +6
& 100 & 96 & -4 \\

Orange juice
& 44 & 76 & +32
& 66 & 78 & +12
& 100 & 98 & -2
& 74 & 82 & +8
& 94 & 98 & +4 \\

\midrule

Average
& 28.2 & 57.2 & +29.0
& 66.4 & 76.4 & +10.0
& 98.0 & 98.6 & +0.6
& 77.8 & 87.8 & +10.0
& 94.0 & 96.0 & +2.0 \\
\bottomrule
\end{tabular*}
\caption{
\textbf{Per-task success rate (\%) on LIBERO-Object.}
Each row picks up the named object and places it in the basket. \emph{Base} denotes the corresponding downstream VLA policy, \method{} denotes its A4A-instantiated variant, and $\Delta$ is the performance difference between them.
}
\label{tab:libero_object_pertask}
\vspace{-3mm}
\end{table*}

\paragraph{Benchmarks.}
We evaluate downstream policies in simulation and the real world and report task success under each benchmark's task-specific success criterion. Because our 4D affordance corpus focuses on object interactions rather than variations in scene layout or task goal, we use the ten LIBERO-Object~\cite{liu2023libero} tasks for the cross-paradigm VLA evaluation. Each policy is trained on 500 demonstrations, 50 per task, and evaluated on 50 episodes per task using the same official initial states. We additionally evaluate four RLBench~\cite{james2020rlbench} tasks---meat off grill, sweep to dustpan, turn tap, and slide block to target---to compare our RGB-based policy with methods that use explicit 3D or 4D observations. Each RLBench policy is trained on 100 demonstrations per variation and evaluated on 25 rollouts per task.

For real-world evaluation, we use an AgileX Piper arm on placing a cup on a mug, opening a microwave, cutting a peach, and three independently trained cooking skills: opening the pot, pouring rice, and pouring water. We collect 100 teleoperated demonstrations per skill using head- and wrist-mounted Intel RealSense D435 cameras and evaluate each policy on 10 rollouts per skill. In every paired comparison, the use of \method{} initialization is the only variable, while the downstream demonstrations, finetuning settings, observations, and evaluation episodes are fixed.

\paragraph{Baselines.}
On LIBERO-Object, we compare five representative VLA policies that cover five commonly used action paradigms. Octo~\cite{team2024octo} is a diffusion-based policy. OpenVLA~\cite{kim2024openvla} is a discrete autoregressive policy. OpenVLA-OFT~\cite{kim2025fine} is a continuous regression policy. $\pi_0$~\cite{black2024pi_0} is an end-to-end flow-matching policy. $\pi_{0.5}$~\cite{black2025pi05} is a hierarchical flow-matching policy. For OpenVLA and OpenVLA-OFT we evaluate their officially released LIBERO checkpoints, for $\pi_0$ and $\pi_{0.5}$ we evaluate the LeRobot~\cite{cadene2024lerobot} ports, and for Octo we fine-tune the small Octo backbone ourselves. 
On RLBench, we compare against representative baselines evaluated on the same tasks, including Image-BC (ViT)~\cite{shridhar2023perceiver}, C2FARM-BC~\cite{james2022coarse}, ManiGaussian~\cite{lu2024manigaussian}, and ARM4R~\cite{niu2025pre}. For ARM4R, we use its released stage-one weights for downstream finetuning, while the remaining results are quoted from prior work.

\paragraph{Implementation.}
For LIBERO-Object, each \method{} variant follows the common affordance-to-action transfer mechanism described in Sec.~\ref{sec:affordance_transfer} while retaining the native action-generation paradigm of its base policy. Model-specific implementations are provided in Appendix A. On RLBench, our policy uses a Qwen3.5-4B~\cite{team2026qwen3} backbone whose causal Transformer serves as the shared latent network. A point projector is used during the 4D stage and an action projector during robot finetuning; both are implemented as two-layer MLPs. We use a context length of 16 following ARM4R, with the full architecture described in Appendix A.6. All models are trained and evaluated on a single host equipped with 8 $\times$ NVIDIA A800 GPUs.




\begin{table}[t]
\centering
\footnotesize
\setlength{\tabcolsep}{5pt}
\begin{tabular}{@{}lccccc@{}}
\toprule
Method & Meat & Sweep & Turn Tap & Slide & Avg. \\
\midrule
Image-BC (ViT) & 0.0 & 0.0 & 16.0 & 0.0 & 4.0 \\
C2FARM-BC & 20.0 & 0.0 & \textbf{68.0} & 16.0 & 26.0 \\
ManiGaussian & 60.0 & \textbf{64.0} & 56.0 & 24.0 & 51.0 \\
ARM4R & 68.0 & 48.0 & 28.0 & 20.0 & 41.0 \\
\midrule
Ours w/o pretrain & 56.0 & 44.0 & 44.0 & 12.0 & 39.0 \\
Ours w/ pretrain & \textbf{88.0} & 60.0 & \textbf{68.0} & \textbf{28.0} & \textbf{61.0} \\
$\Delta$ & +32.0 & +16.0 & +24.0 & +16.0 & +22.0 \\
\bottomrule
\end{tabular}
\caption{
\textbf{Success rate (\%) on RLBench.} Meat, Sweep, Turn Tap, and Slide denote meat off grill, sweep to dustpan, turn tap, and slide block to target. Avg. is the mean success rate over the four tasks. $\Delta$ is the gain from 4D affordance pretraining, and comparison methods are quoted from prior work.
}
\label{tab:rlbench}
\vspace{-3mm}
\end{table}

\subsection{Experiment Analysis}

\begin{table*}[t]
\centering
\small
\renewcommand{\arraystretch}{1.15}
\begin{tabular*}{\textwidth}{@{\extracolsep{\fill}}lccc ccc c@{}}
\toprule
& & & & \multicolumn{3}{c}{Cook Rice} & \\
\cmidrule(lr){5-7}
Method & Microwave & Cup & Peach & Pot & Rice & Water & Avg. \\
\midrule
Octo~\cite{team2024octo} & 7/10 & 3/10 & 3/10 & 2/10 & 4/10 & 3/10 & 36.7\% \\
\quad + \method{} & 8/10 & 5/10 & 6/10 & 5/10 & 6/10 & 6/10 & 60.0\% \\
OpenVLA~\cite{kim2024openvla} & 6/10 & 6/10 & 4/10 & 3/10 & 6/10 & 4/10 & 48.3\% \\
\quad + \method{} & 9/10 & 6/10 & 7/10 & 5/10 & 7/10 & 7/10 & 68.3\% \\
OpenVLA-OFT~\cite{kim2025fine} & 9/10 & 7/10 & 7/10 & 5/10 & 8/10 & 7/10 & 71.7\% \\
\quad + \method{} & 9/10 & 8/10 & 9/10 & 6/10 & 8/10 & 8/10 & 80.0\% \\
$\pi_{0.5}$~\cite{black2025pi05} & 9/10 & 8/10 & 7/10 & 5/10 & 7/10 & 6/10 & 70.0\% \\
\quad + \method{} & 9/10 & 9/10 & 9/10 & 7/10 & 8/10 & 7/10 & 81.7\% \\
\bottomrule
\end{tabular*}
\caption{
\textbf{Success on the physical Piper arm.} Cells are successful trials out of 10. Microwave, Cup, and Peach denote open the microwave, place the cup on the mug, and cut the peach. Cook rice is reported as three subtasks, Pot (open the pot), Rice (pour rice), and Water (pour water). Avg. is the mean success rate over the six tasks. +\method{} denotes the corresponding policy trained through our affordance-to-action framework.
}
\label{tab:piper_real}
\vspace{-5mm}
\end{table*}

\paragraph{Effectiveness Across VLA Paradigms.}
Table~\ref{tab:libero_object_pertask} reports results for five representative VLA policy families with distinct action-generation paradigms, covering diffusion, discrete autoregressive generation, continuous regression, end-to-end flow matching, and hierarchical flow matching. \method{} improves the average success rate for every policy, by $29.0$ percentage points for Octo, $10.0$ for OpenVLA, $0.6$ for OpenVLA-OFT, $10.0$ for $\pi_0$, and $2.0$ for $\pi_{0.5}$. These gains across different architectures, action representations, and learning objectives show that the effectiveness of \method{} is not tied to a particular VLA formulation. By transferring task-conditioned interaction geometry from human demonstrations into each policy's native action-generation process, \method{} provides a complementary prior for downstream robot control. The smaller gains for OpenVLA-OFT and $\pi_{0.5}$ are consistent with their high baseline success rates of $98.0\%$ and $94.0\%$, respectively, which leave limited headroom for further improvement. Nevertheless, \method{} still yields modest gains for both policies. Overall, these results demonstrate that action-oriented 4D affordances provide a broadly applicable cross-embodiment prior for improving robot control across heterogeneous VLA formulations.

\paragraph{Comparison with Geometry-Aware Policies.}
Table~\ref{tab:rlbench} shows that action-oriented 4D affordance supervision improves all four RLBench tasks, raising the average success rate of our RGB policy from $39.0\%$ to $61.0\%$ and achieving the highest average among the compared methods. The gains are especially pronounced on meat off grill and turn tap, suggesting that the learned representation benefits tasks requiring precise spatial reasoning and contact-dependent motion. 
Notably, without using explicit 3D observations during downstream execution, the resulting policy performs best on meat off grill and slide block to target, while matching the best result on turn tap. These results show that \method{} can internalize interaction-relevant 4D geometry from human demonstrations and provide an effective geometric prior for RGB-based manipulation.


\paragraph{Effect of Pretraining Supervision under a Matched Architecture.}
\begin{figure}[t]
\centering
\includegraphics[width=\columnwidth]{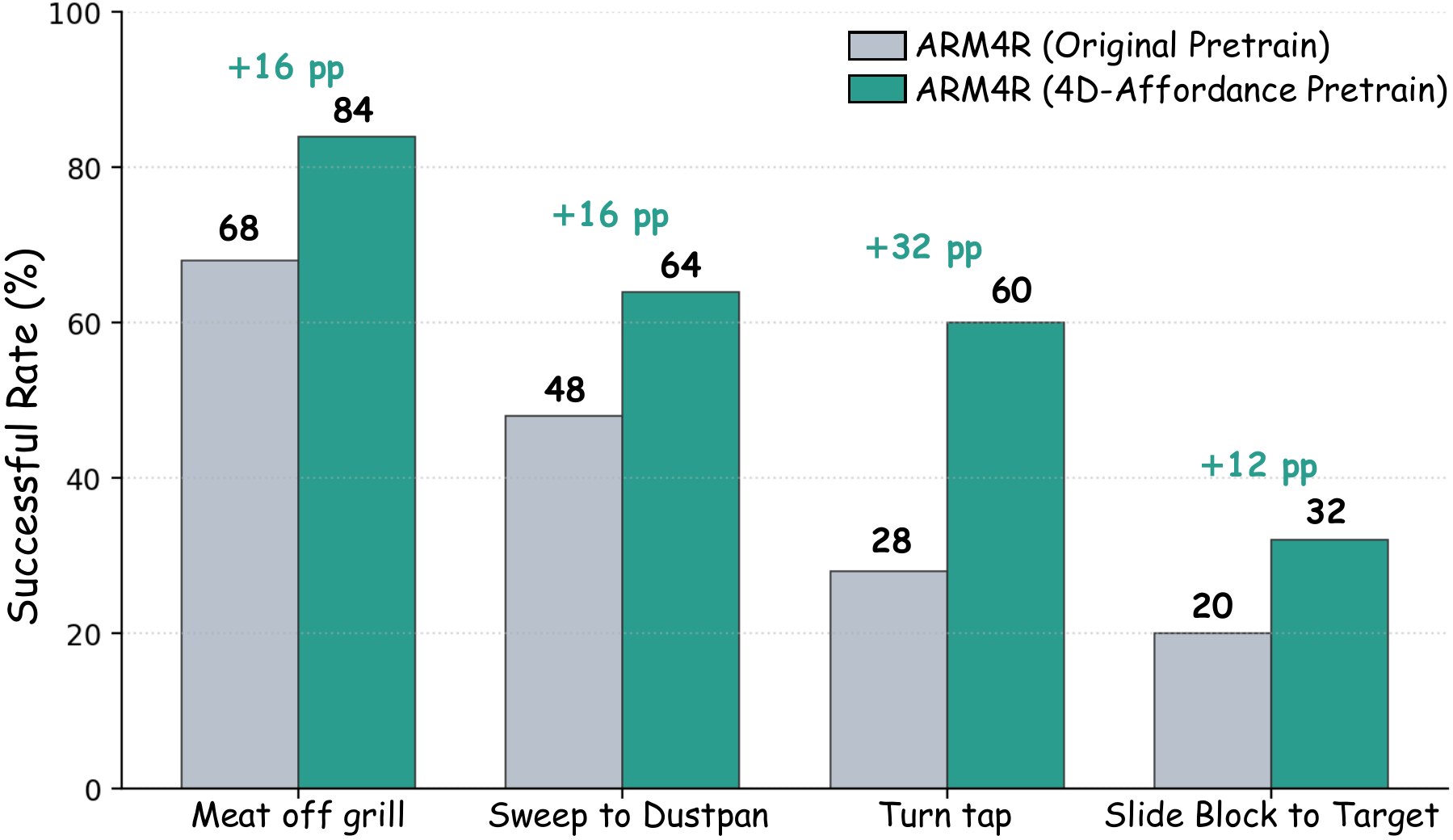}
\caption{\textbf{Architecture-controlled comparison.} Both variants use the same ARM4R architecture, robot demonstrations, finetuning recipe, and evaluation protocol. Gray uses the released initialization learned from generic scene-wide 4D point tracks; green uses the same architecture pretrained on our action-oriented 4D affordance dataset.}
\label{fig:dataset_improvement}
\vspace{-5mm}
\end{figure}

To isolate the effect of stage-one pretraining data and supervision, we compare two ARM4R variants with the same backbone, robot demonstrations, finetuning procedure, and evaluation protocol. The baseline uses ARM4R's released initialization learned from generic scene-wide 4D point tracks, whereas the other variant pretrains the same architecture from scratch on our action-oriented 4D affordance dataset. As shown in Figure~\ref{fig:dataset_improvement}, our supervision improves all four RLBench tasks, increasing the average success rate from $41.0\%$ to $60.0\%$, with turn tap increasing from $28.0\%$ to $60.0\%$. Under this architecture-controlled setting, the results demonstrate that action-oriented 4D affordance supervision provides a more effective transferable prior for robot control than generic scene-wide 4D motion.


\paragraph{Real-World Results.}

Table~\ref{tab:piper_real} evaluates \method{} on physical manipulation tasks using four representative VLA policies. Incorporating action-oriented 4D affordances improves the average success rate for each policy, increasing the mean success rate across all policies from $56.7\%$ to $72.5\%$. Clear improvements are observed on pouring rice and pouring water, where successful execution requires coordinated object translation and substantial orientation changes. These results highlight an important advantage of learning from human demonstrations: the extracted 4D representation captures not only where the interaction occurs, but also how the interaction-relevant object geometry should evolve throughout the operation, including the rotational motion required for pouring. Such operational information complements robot demonstrations and improves the policy's ability to execute contact-rich manipulation in physical environments. Overall, the real-world results demonstrate that action-oriented 4D affordances provide an effective manipulation prior for physical robot control. Representative executions are shown in Figure~\ref{fig:piper_snapshots}.

\section{Limitations}

The current instantiation of \method{} remains limited on tasks involving large non-rigid deformation, severe slippage or repeated contact switching, and complex bimanual coordination. For example, multi-turn bottle-cap opening requires repeated release, regrasping, and contact re-establishment, which are not modeled by the current transfer formulation. In addition to extending \method{} to these more complex interactions, a complementary direction is to introduce a further training stage using 4D affordances constructed from robot motion. Such robot-domain training is not considered in our current setting but may further improve downstream control performance.


\section{Conclusion}

We presented \method{}, a cross-embodiment transfer framework that represents human manipulation as action-oriented 4D affordances---the language-conditioned future trajectories of interaction-relevant 3D points. Motivated by their short-horizon geometric correspondence with robot end-effector motion, we construct a large-scale 4D affordance dataset and use point-trajectory prediction to initialize VLA policies before adapting them to their native state and action interfaces. Experiments across five VLA families, simulated benchmarks, and real-world tasks show consistent downstream improvements, while the architecture-controlled comparison further validates the effectiveness of interaction-centric 4D supervision. These results establish action-oriented 4D affordances as an effective cross-embodiment representation for transferring manipulation knowledge from human demonstrations to robot control.

\bibliography{aaai2027}
\appendix
\twocolumn[%
  \begin{center}
    {\Large\bfseries Supplementary Material for\\[0.5em]
    $\mathcal{A}\textit{4}\mathcal{A}$:
    Cross-Embodiment Transfer of Action-Oriented 4D Affordances\\
    from Human Demonstrations}
    \vspace{1em}
  \end{center}
]
\section{Actionable 4D Affordance Corpus}
\label{app:affordance_data}

\subsection{Corpus Composition and Affordance Coverage}

Our actionable 4D affordance corpus combines human--object interaction data from HOI4D~\cite{liu2022hoi4d} and EPIC-KITCHENS~\cite{damen2020epic} with complementary RGB-D demonstrations collected in-house. The resulting corpus contains over 80K interaction clips, of which approximately 30K are obtained from our in-house recordings. The public datasets provide naturally occurring interactions across diverse objects, environments, viewpoints, and human behaviors, whereas the in-house subset is designed to expand the coverage of manipulation motions that are less frequent or less geometrically complete in existing data.

The public portion contains common affordance categories such as opening, closing, pushing, pulling, and pressing. We supplement these interactions with pouring, cutting, hanging, sweeping, and lid-removal demonstrations, together with additional articulated-object, tool-mediated, and object-to-object manipulation examples. The resulting corpus therefore spans translation-dominant motion, rotation-dominant motion, constrained articulated motion, and interactions in which one object is used to affect another.

\subsection{Annotation and Trajectory Reconstruction}

We convert heterogeneous demonstrations into a common interaction-centric point-trajectory representation. Task semantics and interaction entities are obtained from dataset annotations when available and otherwise normalized using vision--language parsing. For each demonstration, we localize the manipulated object, tool, or contacted object part and obtain a verified segmentation mask for the corresponding interaction-relevant region. Query points are sampled within this region and tracked through the demonstration while preserving point identity.

For data with reliable depth, object poses, or 3D annotations, the query trajectories are reconstructed directly in 3D. Otherwise, image-space tracks are lifted using estimated depth, camera intrinsics, and camera-motion compensation. The resulting trajectories are transformed into a clip-level reference frame. We remove trajectories affected by invalid depth, prolonged occlusion, unstable tracking, abrupt geometric discontinuities, isolated point clusters, or floating outliers.

For the in-house RGB-D subset, raw depth is refined using LingBot-Depth~\cite{tan2026masked}. GroundingDINO~\cite{liu2024grounding} and SAM~2~\cite{ravi2025sam} provide task-conditioned localization and segmentation, while CoTracker3~\cite{karaev2025cotracker3} preserves query identities across frames. Back-projecting the verified tracks through calibrated RGB-D observations produces metric 4D trajectories that retain both translational and rotational changes of the interaction-relevant geometry.

\begin{figure*}[!t]
\centering
\includegraphics[scale=0.85]{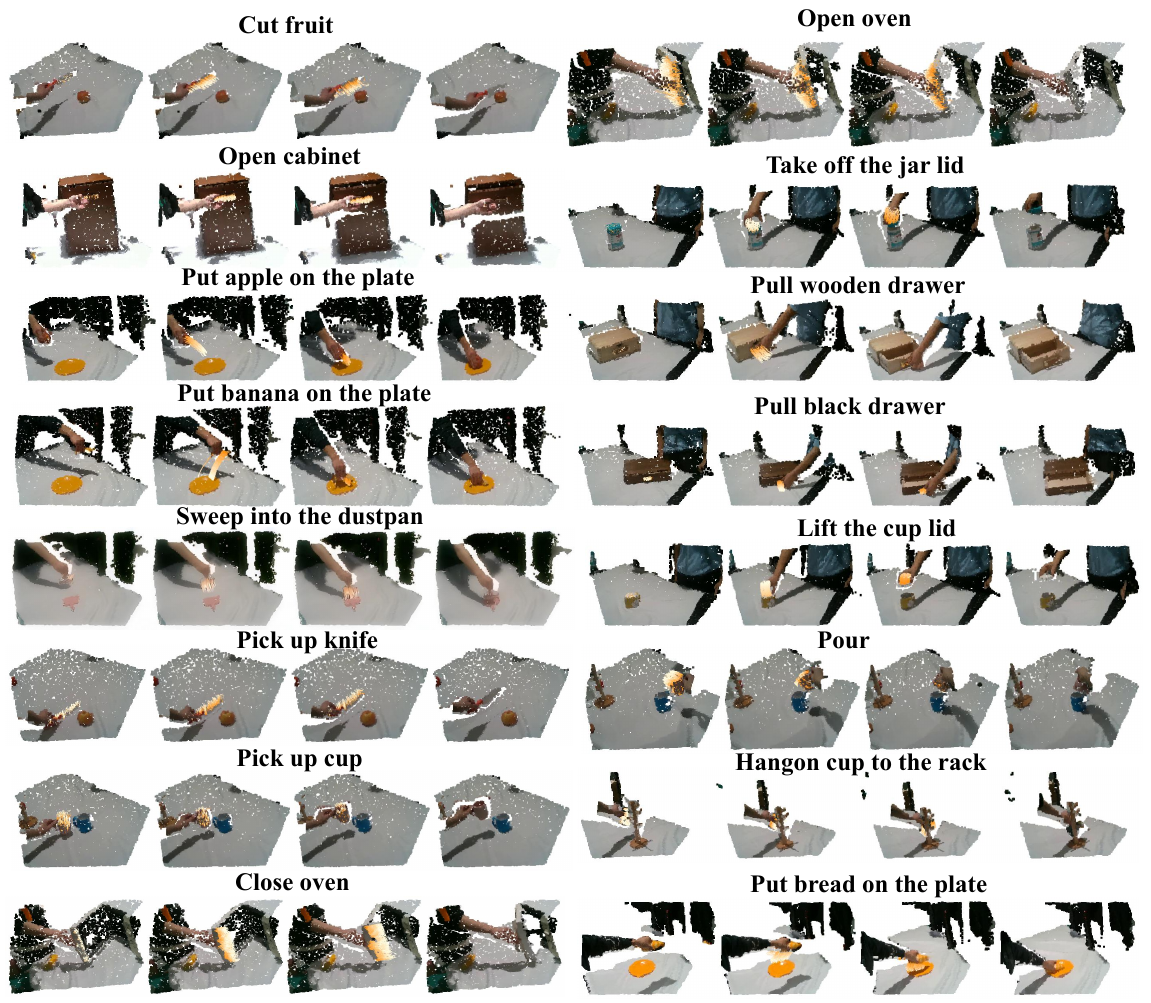}
\caption{
\textbf{Visualization of actionable 4D affordances.}
The colored point flows represent the displacement of interaction-relevant 3D points from the current timestep to the next, with the color gradient indicating the temporal evolution of the motion.
}
\label{fig:affordance_data_visualization}
\end{figure*}

\subsection{Qualitative Visualization}

Figure~\ref{fig:affordance_data_visualization} visualizes representative examples from the constructed corpus. The examples include object placement and pickup, articulated opening and closing, drawer pulling, lid removal, pouring, cutting, sweeping, and hanging. Together, they illustrate the diversity of geometric transitions captured by the corpus, ranging from short translations to large orientation changes and tool-mediated interactions.

\subsection{Data Independence}
The human pretraining corpus and the downstream robot datasets are
collected independently. The human demonstrations are captured with
randomly varying viewpoints, object instances, and scenes, while the
downstream robot demonstrations and evaluations use a different
embodiment, separate environments, and a distinct set of physical
objects. In particular, none of the objects used for real-robot training
or evaluation appear in the in-house human demonstrations. The two
stages therefore share no object instances, scenes, or paired
demonstrations, and overlap only at the level of high-level manipulation
categories, such as opening, pouring, cutting, and sweeping. This setup
evaluates whether manipulation knowledge transfers across embodiments
rather than through instance- or scene-level matching.

\begin{table}[t]
\centering
\footnotesize
\setlength{\tabcolsep}{4pt}
\begin{tabular}{p{0.19\columnwidth}p{0.53\columnwidth}p{0.16\columnwidth}}
\toprule
Subset & Primary affordance coverage & Scale \\
\midrule
Existing datasets
& Open, close, pickup, place, push, pull, press, and naturally occurring object interactions
& Public data \\
In-house RGB-D
& Pour, cut, hang, sweep, lid removal, and additional articulated, tool-mediated, and object-to-object interactions
& $\sim$30K clips \\
\midrule
Combined corpus
& More than ten affordance categories with diverse translational and rotational interaction patterns
& $>$80K clips \\
\bottomrule
\end{tabular}
\caption{
Composition of the actionable 4D affordance corpus. Existing human-interaction datasets provide natural visual diversity, while the in-house RGB-D subset expands the affordance taxonomy and supplies geometrically consistent supervision for underrepresented manipulation motions.
}
\label{tab:affordance_corpus}
\end{table}
\section{A4A across Five VLA Action Paradigms}
\label{app:paradigms}

This section describes how \method{} is instantiated on five VLA policy families while preserving their native action-generation mechanisms. We follow the two-stage training scheme introduced
in Sec.~3.3. OpenVLA uses discrete autoregressive generation, OpenVLA-OFT uses continuous regression, Octo uses diffusion, $\pi_0$ uses end-to-end flow matching, and $\pi_{0.5}$ uses hierarchical flow matching. In each case, the 4D prediction objective follows the same objective family as the native robot-action objective. This paradigm-matched design allows the two stages to share the principal prediction module while using stage-appropriate state and output representations.

\subsection{Unified Affordance-to-Action Transfer}

\begin{figure*}[!t]
    \centering
    \includegraphics[scale=0.85]
    {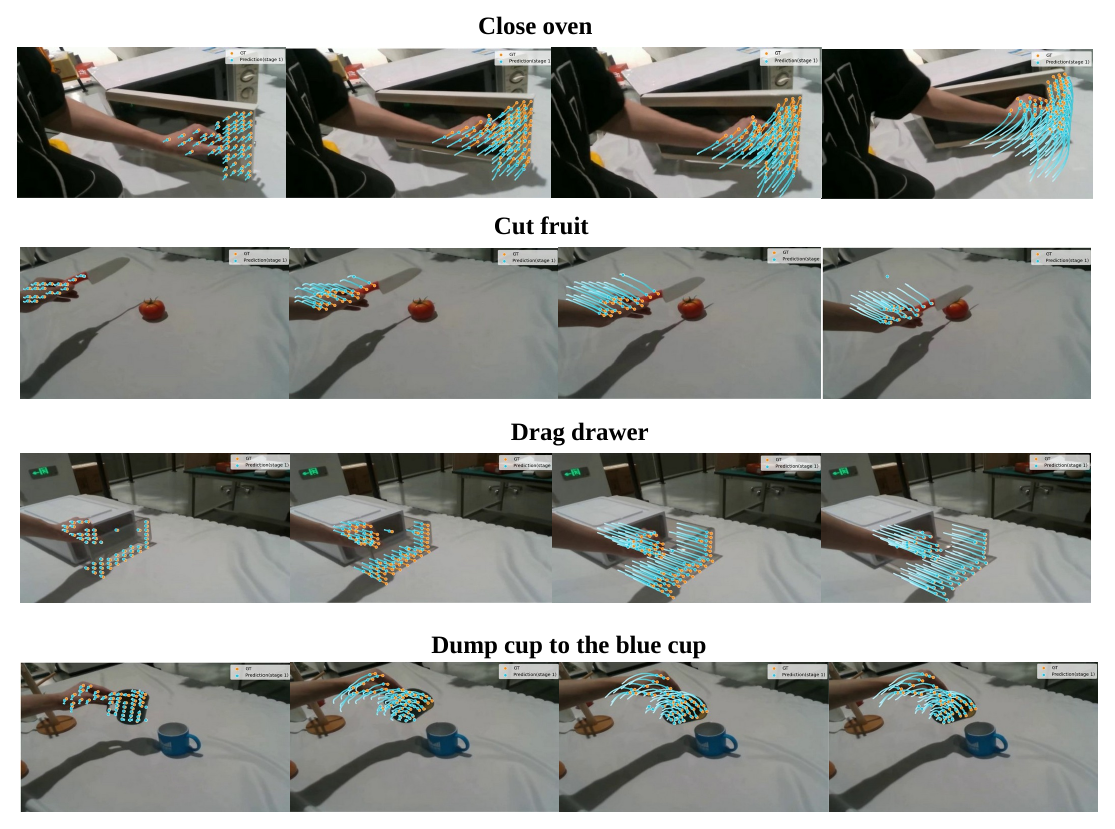}
    \caption{
    \textbf{Qualitative results of stage-one actionable 4D
    affordance prediction.}
    We compare the predicted and ground-truth point-motion targets
    across four representative interactions: closing
    an oven, cutting fruit, dragging a drawer, and pouring from one cup into another.
    }
    \label{fig:affordance_prediction}
\end{figure*}

\paragraph{Point-motion target.}
Let $\mathbf{p}_i^t\in\mathbb{R}^3$ denote the position of query point $i$ at the final context frame. Given $C$ context frames, we predict a short-horizon trajectory residual relative to constant-velocity extrapolation:
\begin{equation}
\begin{aligned}
\mathbf{v}_i^t
&=
\mathbf{p}_i^t-\mathbf{p}_i^{t-1},\\
\mathbf{r}_{i,h}
&=
\left(\mathbf{p}_i^{t+h}-\mathbf{p}_i^t\right)
-h\mathbf{v}_i^t,\\
\widehat{\mathbf{p}}_i^{t+h}
&=
\mathbf{p}_i^t+h\mathbf{v}_i^t+\widehat{\mathbf{r}}_{i,h},
\qquad h=1,\ldots,H.
\end{aligned}
\label{eq:appendix_cv_residual}
\end{equation}
For each training target, the cumulative displacement of each query point
over the next $H$ frames is represented as a trajectory residual relative
to constant-velocity extrapolation. Residuals are normalized per coordinate using statistics computed from the pretraining corpus, and invalid or occluded tracks are excluded from the loss. This parameterization captures the near-future motion induced by an interaction without introducing a long-horizon forecasting objective, while assigning zero residual to locally constant-velocity motion. Figure~\ref{fig:affordance_prediction} provides qualitative
comparisons between the predicted and ground-truth point-motion
targets of the stage-one affordance model.
\begin{table*}[t]
\centering
\footnotesize
\setlength{\tabcolsep}{5pt}
\resizebox{0.85\textwidth}{!}{%
\begin{tabular}{lllll}
\toprule
Base policy & Native paradigm & Transferred module & 4D pretraining objective & Robot-action objective \\
\midrule
OpenVLA
& Discrete autoregressive
& Llama-2-7B decoder and output head
& Point-token cross-entropy
& Action-token cross-entropy \\
OpenVLA-OFT
& Continuous regression
& Llama-2-7B decoder
& $L_1$/MSE point regression
& $L_1$/MSE action regression \\
Octo
& DDPM diffusion
& Block-causal Transformer
& Point-residual denoising
& Action denoising \\
$\pi_0$
& End-to-end flow matching
& Low-level action expert
& Flow matching on point residuals
& Flow matching on actions \\
$\pi_{0.5}$
& Hierarchical flow matching
& Low-level action expert
& Flow matching on point residuals
& Flow matching on actions \\
\bottomrule
\end{tabular}
}
\caption{
Instantiation of \method{} across five VLA action paradigms. Each policy retains its native prediction pathway and objective family while replacing the stage-specific state and output interfaces.
}
\label{tab:paradigm_pretrain}
\end{table*}

\begin{table*}[t]
\centering
\footnotesize
\setlength{\tabcolsep}{5pt}
\resizebox{0.85\textwidth}{!}{%
\begin{tabular}{llllcc}
\toprule
Policy & Image input & Camera views & Action formulation & Execution horizon & Inference steps \\
\midrule
OpenVLA
& $224$ px, center crop
& Third-person
& Autoregressive tokens
& $1$
& --- \\
OpenVLA-OFT
& $224$ px, center crop
& Third-person + wrist
& Continuous regression
& $8$
& --- \\
Octo
& $256/128$ px
& Third-person + wrist
& DDPM diffusion
& $4$
& Native \\
$\pi_0$
& $256\!\rightarrow\!224$ px, padded
& Third-person + wrist
& Flow matching
& $50$
& $10$ \\
$\pi_{0.5}$
& $256\!\rightarrow\!224$ px, padded
& Third-person + wrist
& Flow matching
& $50$
& $10$ \\
\bottomrule
\end{tabular}
}
\caption{
Policy-specific inference settings on LIBERO-Object. Execution horizon is the number of actions executed per policy query; inference steps apply only to iterative diffusion or flow-matching policies.
}
\label{tab:eval_hparams}
\end{table*}

\paragraph{Paradigm-matched objective.}
For a base policy $k$, let $\mathcal{J}_k$ denote its native prediction objective. The corresponding 4D pretraining loss is
\begin{equation}
\mathcal{L}_{\mathrm{4D}}^{(k)}
=
\mathcal{J}_k
\left(
\widehat{\mathbf{R}},
\mathbf{R};
\mathbf{M}
\right),
\label{eq:appendix_matched_objective}
\end{equation}
where $\mathbf{R}$ collects the short-horizon point-trajectory residuals and $\mathbf{M}$ masks invalid trajectories. The form of $\mathcal{J}_k$ is retained across pretraining and robot finetuning: token cross-entropy for OpenVLA, continuous regression for OpenVLA-OFT, DDPM denoising for Octo, and flow matching for $\pi_0$ and $\pi_{0.5}$. Only the state interface and prediction space are adapted to the corresponding stage.

\subsection{Discrete Autoregressive Generation: OpenVLA}

OpenVLA~\cite{kim2024openvla} represents continuous robot actions as discrete tokens and generates them autoregressively with its language-model decoder. For 4D pretraining, each coordinate of the normalized point residual is quantized using the same discretization principle as the action representation. Visual and language tokens are augmented with embeddings of the current query-point state, and the decoder predicts the future point-residual tokens under cross-entropy supervision.

Robot finetuning restores OpenVLA's native action-token targets and removes the point-specific input interface. The language-model decoder and output head learned during 4D prediction initialize the downstream policy. Consequently, both stages use the same autoregressive prediction pathway; they differ only in whether the generated tokens represent point motion or embodiment-specific robot actions.

\subsection{Continuous Regression: OpenVLA-OFT}

OpenVLA-OFT~\cite{kim2025fine} replaces discrete autoregressive action generation with parallel continuous regression. During 4D pretraining, a lightweight point-state projector embeds the query-point coordinates, and the language-model decoder predicts the continuous point residuals using the same $L_1$ or MSE objective family employed for robot actions. This preserves the continuous output geometry of the native policy.

During robot finetuning, the point-state projector is replaced by the native proprioceptive interface, and the prediction target is changed from point residuals to action chunks. The decoder parameters learned from interaction-centric point prediction are retained and jointly optimized with the downstream action interface. Thus, the transfer occurs through the shared decoder rather than through an auxiliary temporal model.

\subsection{Diffusion Generation: Octo}

Octo~\cite{team2024octo} generates action chunks with a diffusion objective conditioned on visual observations and language. We retain its block-causal Transformer and use the current query-point state as an additional observation stream. The future point residuals replace the robot-action chunk as the diffusion target, while the native noise-conditioning mechanism and DDPM denoising objective are preserved.

Robot finetuning restores the original action dimensionality and action-conditioning interface. The visual--language conditioning modules and block-causal Transformer are initialized from 4D pretraining, while lightweight stage-specific projections accommodate the different point and action dimensions. The transferred representation is therefore learned through the same denoising process used for downstream action generation.

\subsection{Flow Matching: $\pi_0$ and $\pi_{0.5}$}
\label{app:flow}

$\pi_0$~\cite{black2024pi_0} generates low-level action chunks directly through flow matching. $\pi_{0.5}$~\cite{black2025pi05} retains a related low-level flow-matching action expert within a hierarchical policy that additionally predicts a high-level semantic representation. A4A transfers through the low-level action expert shared with robot-action generation; the pathway for high-level semantic subtask prediction is kept frozen.

For 4D pretraining, the point residual $\mathbf{r}$ is treated as the clean endpoint of the flow process. At flow time $\tau\in[0,1]$, we construct
\begin{equation}
\mathbf{x}_{\tau}
=
(1-\tau)\mathbf{r}
+\tau\boldsymbol{\epsilon},
\qquad
\mathbf{u}_{\tau}
=
\boldsymbol{\epsilon}-\mathbf{r},
\label{eq:appendix_point_flow}
\end{equation}
where $\boldsymbol{\epsilon}$ is Gaussian noise. A lightweight point projector maps the noised residuals into the action-expert input space, and the model predicts the velocity field $\mathbf{u}_{\tau}$ under mean-squared error. Visual and language features condition the point-flow prediction through the policy's native multimodal pathway.

During robot finetuning, the point-specific projections are replaced by the native state and action projections, and the target becomes the robot action chunk. The low-level action expert is initialized from 4D pretraining and retains the native flow-time conditioning. For $\pi_{0.5}$, this transfer is confined to the low-level action-generation pathway, preserving the hierarchical organization of the base policy.

\subsection{RLBench Policy Instantiation}
\label{app:rlbench_architecture}

Our RLBench policy instantiates the same transfer principle with a Qwen3.5-4B backbone. Its causal Transformer, with hidden width $2048$, serves as the shared prediction module. During 4D pretraining, a two-layer point projector maps the query-point state into the Transformer input space, and the model predicts the point residual defined in Eq.~\eqref{eq:appendix_cv_residual}. During robot finetuning, the point interface is replaced by a two-layer proprioceptive projector and a two-layer action head, while the causal Transformer is retained and jointly finetuned. Both stages use a context length of $16$, matching the temporal context used in the architecture-controlled comparison with ARM4R.

\subsection{LIBERO Evaluation Protocol}
\label{app:libero_protocol}

All five policy families are evaluated on LIBERO-Object using a common rollout protocol while retaining each policy's native
observation and inference configuration. Each task is evaluated for $50$ rollouts from the same official initial states, using the benchmark's task-specific success predicate and a maximum horizon of $280$ environment steps. Actions are converted to the benchmark control space using the normalization statistics associated with each base policy. Table~\ref{tab:eval_hparams} summarizes the policy-specific observation and inference settings retained from the corresponding native implementations.

OpenVLA-OFT additionally receives proprioceptive state and is adapted with rank-$32$ LoRA. The $\pi_0$ and $\pi_{0.5}$ policies use a single observation frame and preserve their native action-chunk horizon. These policy-specific settings are held fixed between the corresponding base and A4A-instantiated evaluations.

\end{document}